\documentclass[10pt,a4paper]{article}
\usepackage[left=3cm,right=3cm]{geometry}
\usepackage[english]{babel}

\usepackage{amsmath,amsfonts,bm}

\def\eqref#1{equation~\ref{#1}}

\def\1{\bm{1}}

\def\vu{{\bm{u}}}

\DeclareMathAlphabet{\mathsfit}{\encodingdefault}{\sfdefault}{m}{sl}
\SetMathAlphabet{\mathsfit}{bold}{\encodingdefault}{\sfdefault}{bx}{n}

\newcommand{\R}{\mathbb{R}}

\DeclareMathOperator*{\argmin}{arg\,min}

\usepackage{graphicx} 
\usepackage{hyperref}
\usepackage{url}
\usepackage{tikz}
\usepackage{amssymb}
\usepackage{amsfonts}
\usepackage{amsmath}
\usepackage{stmaryrd}
\usepackage{amsthm}

\newtheorem{proposition}{Proposition}

\newtheorem{corollary}{Corollary}

\newtheorem{theorem}{Theorem}
\newcommand{\N}{\mathbb{N}}
\tikzstyle{mynode}=[thick,draw=blue,fill=blue!20,circle,minimum size=22]
\tikzstyle{mynodebis}=[thick,draw=blue,fill=blue!25,circle,minimum size=22]

\tikzstyle{mynodeinput}=[thick,draw=red,fill=red!20,circle,minimum size=22]
\tikzstyle{mynodeoutput}=[thick,draw=green,fill=green!20,circle,minimum size=22]

\title{Optimization Insights into Deep Diagonal Linear Networks}

\author{Hippolyte Labarri\`ere\footnote{MaLGa, DIBRIS, Universit\`a degli Studi di
Genova, Genoa, Italy.} \and Cesare Molinari\footnote{MaLGa, DIMA, Universit\`a degli Studi di
Genova, Genoa, Italy.} \and Lorenzo Rosasco \footnotemark[1] \footnotemark[2] \footnote{Center for Brains, Minds and Machines, MIT, Cambridge, MA, United States of
America.} \footnote{Istituto Italiano di Tecnologia, Genoa, Italy.} \and Cristian Vega\footnote{Instituto de Alta investigaci\`on (IAI), Universidad de Tarapac\`a, Arica, Chile.} \and Silvia Villa\footnotemark[2]}
\date{}
\begin{document}

\maketitle

\begin{abstract}
Gradient-based methods successfully train highly overparameterized models in practice, even though the associated optimization problems are markedly nonconvex. Understanding the mechanisms that make such methods effective has become a central problem in modern optimization. To investigate this question in a tractable setting, we study Deep Diagonal Linear Networks. These are  multilayer architectures with a reparameterization that preserves convexity in the effective parameter,  while inducing a nontrivial geometry in the optimization landscape. Under mild initialization conditions, we show that gradient flow on the layer parameters induces a mirror-flow dynamic in the effective parameter space. This structural insight yields explicit convergence guarantees, including exponential decay of the loss under a Polyak–\L{}ojasiewicz condition, and clarifies how the parametrization and initialization scale govern the training speed. Overall, our results demonstrate that deep diagonal over parameterizations, despite their apparent complexity, can endow standard gradient methods with well-behaved and interpretable optimization dynamics.

\end{abstract}

\section{Introduction}

In recent years, the application of deep neural networks has transformed the field of machine learning, achieving outstanding results in tasks involving complex data such as images and natural language. These models are typically trained successfully using gradient-based methods, most notably stochastic gradient descent,  consistently reaching zero empirical loss across various benchmarks. This empirical success is particularly striking given the highly nonconvex nature of the underlying optimization problems, where classical theory would provide little/no guarantees. Understanding why such simple optimization procedures are so effective in training overparameterized models has therefore become a central research question.

\begin{figure}[ht]
\begin{center}
\begin{tikzpicture}[x=1.5cm,y=0.8cm]
 \newcounter{counter}
 \foreach \N [count=\lay,remember={\N as \Nprev (initially 0);}]
        in {4,4,4,1}{ 
  \stepcounter{counter}
  \foreach \i [evaluate={\y=\N/2-\i; \x=\lay; \prev=int(\lay-1);}]
         in {1,...,\N}{ 
   \ifnum\value{counter}=1
    \node[mynodeinput] (N\lay-\i) at (\x,\y) {$x^i_\i$};
   \fi
   \ifnum\value{counter}=2
    \node[mynode] (N\lay-\i) at (\x,\y) {$u_\i$};
   \fi
   \ifnum\value{counter}=3
    \node[mynodebis] (N\lay-\i) at (\x,\y) {$v_\i$};
   \fi
   \ifnum\value{counter}=4
    \node[mynodeoutput] (N\lay-\i) at (\x,\y) {$f_\theta(x^i)$};
   \fi
   \ifnum\Nprev>0 
    \ifnum\N>1
      \draw[thick] (N\prev-\i) -- (N\lay-\i);
    \fi
    \ifnum\N<2
    \foreach \j in {1,...,\Nprev}{ 
     \draw[thick] (N\prev-\j) -- (N\lay-\i);
    }
    \fi
   \fi
  }
 }
\end{tikzpicture}
\end{center}
 \caption{Representation of a Diagonal Linear Network}
 \label{fig:DLN}
\end{figure}
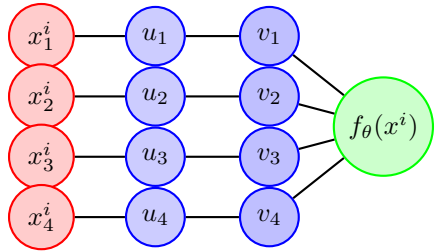

A common approach to understanding optimization in overparameterized models is to analyze simplified settings where the interaction between parameterization and optimization becomes more transparent. Even in these reduced models, the dynamics often remain surprisingly intricate, yet they can shed light on how parameterization shapes convergence behavior.
Within this line of work, Diagonal Linear Networks (DLNs), introduced independently by \cite{woodworth2020kernel} and \cite{moroshko2020implicit}, have emerged as a particularly tractable framework. In shallow DLNs, the network function takes the form $f_\theta(x) = \theta^\top x$ with parameters $\theta = u \odot v$, where $\odot$ denotes the Hadamard product (see Figure \ref{fig:DLN}). Despite their simplicity, DLNs display a variety of nontrivial optimization phenomena. Prior studies have examined how initialization scale, step size, or stochasticity influence convergence and the nature of the obtained solution \cite{azulay2021implicit,pesme2021implicit,mathieu,nacson2022implicit,papazov2024leveraging}. These works highlight, among other insights, that mild overparameterization can already  smooth the loss landscape, and that small versus large initializations give rise to qualitatively different training behaviors, mirroring the so called rich and lazy regimes of \cite{NEURIPS2019_ae614c55}.

In this paper, we extend this line of inquiry to Deep Diagonal Linear Networks, which generalize DLNs to an $L$-layer parametrization. In this model, the effective parameter is given by
\begin{equation*}
    \theta = u^1\odot u^2\odot \dots \odot u^L,
\end{equation*}
where each $u^j \in \R^d$. This formulation offers a closer abstraction of multilayer architectures while preserving analytical tractability, and it allows us to explore how depth influences the optimization landscape and the implicit dynamics induced by training. Note that a simpler version of the $L$-layers case has been studied in \cite{woodworth2020kernel,moroshko2020implicit,chou2023more} considering the simpler parameterization $\theta=u^{\odot L}-v^{\odot L}$.

Our goal is to understand the effect of this reparameterization when the layers weights $(u^j)_{j\in[L]}$ are trained using gradient flow. More precisely, we consider the problem
\begin{equation*}
\min_{\theta\in\R^d}\mathcal{L}(\theta),
\end{equation*}
for some differentiable loss function $\mathcal{L}:\R^{d}\rightarrow\R$, and for some input-output data $X:\R^d\rightarrow\R^n$ and $y\in\R^n$. We optimize $\mathcal{L}$ by applying gradient flow with respect to the variables $(u^j)_{j\in[L]}$. 

Our main contributions are summarized as follows:
\begin{itemize}
\item We provide a mild initialization condition under which gradient flow on $(u^j)_{j\in[L]}$ induces a trajectory in $\theta$ that follows a mirror-flow dynamic.
\item Under the same condition, we establish several convergence guarantees for the original optimization problem. In particular, when the loss function $\mathcal{L}$ satisfies a Polyak-\L{}ojasiewicz condition, the objective function in the $\theta$-space decreases exponentially toward its minimum. Our analysis also indicates that very small initializations lead to slow training, consistent with phenomena observed in related models.
\item We identify quantities that are conserved along the gradient-flow trajectory,  in line with recent works on conservative laws \cite{marcotte2023abide,marcotte2025transformative}.  In particular, this reveals that the dynamics of $\theta$ are governed by those of an abstract layer associated with minimal components.
\end{itemize}

We provide a detailed review of related topics in the literature in the next section.

\subsection{Related works}

\paragraph{Conservation laws} A common approach to studying the behavior of optimization dynamics in neural network models is to identify the conserved quantities that remain invariant during training. Such quantities often provide insight into convergence properties or implicit bias phenomena (see, e.g., \cite{saxe2013exact,du2018algorithmic}). Characterizing these conservation laws is therefore crucial for understanding the overall training dynamics. Recent works have pursued this direction by establishing convergence laws for gradient flow in general neural network models \cite{marcotte2023abide}, momentum-based dynamics \cite{marcotte2024keep}, and discrete-time dynamics \cite{marcotte2025transformative}.

\paragraph{Hadamard parameterization to promote sparsity} Diagonal Linear Networks are closely related to Hadamard parameterization (HP) and it is worth noticing that HP was used before for sparsity recovery. \cite{hoff2017lasso} takes benefit of HP to promote sparsity in a LASSO problem. \cite{vaskevicius2019implicit,zhao2022high} also consider an HP for a least squares problem inducing an implicit bias towards the minimal $L^1$-norm solution, enabling the use of early stopping strategies. In a comparable vein, \cite{amid2020winnowing} exploit a reparameterization $\theta=u^{\odot2}$ for a classification problem and \cite{chou2022non} apply HP to solve non negative least squares. \cite{poon2023smooth} elegantly reformulate a group-LASSO problem using some Hadamard parameterization in order to apply more efficient optimization methods. The geometry induced by HP is also a subject of interest and in this regard, \cite{ouyang2024kurdykalojasiewiczexponenthadamardparametrization} give necessary condition for a reparameterized function to satisfy the Kurdyka-\L{}ojasiewicz property.

\paragraph{Links with Mirror Flow} As shown in \cite{azulay2021implicit,chou2023more}, Diagonal Linear Networks trained with a Gradient Flow (or Gradient Descent) share direct links with Mirror Flow (or Mirror Descent). One of the first appearances of the Mirror Flow dynamic can be attributed to \cite{alvarez2004hessian}. Many years later, a number of studies are presented in the context of understanding the effects of reparameterization. The mirror gradient method having an implicit bias towards specific solutions, \cite{vaskevicius2020statistical} propose early stopping strategies taking advantage of it. \cite{li2022implicit} provide assumptions that are necessary for a reparameterization trained with a Gradient Flow to be equivalent to a Mirror Flow. In their paper, \cite{chou2023induce} adopt an original point of view, seeing reparameterization as a way to enforce some implicit bias related to the corresponding mirror map. In this perspective, the authors give guidelines to define it efficiently.

\paragraph{Matrix factorization} 

A significant step toward understanding the optimization dynamics of overparameterized models is provided by \cite{gunasekar2017implicit}, who analyze gradient-based training in a simple least squares setting,
\[
\mathcal{L}(\Theta) = \|A\Theta - b\|^2, \quad \Theta \in \mathbb{R}^{n \times n}.
\]
By introducing the factorization $\Theta = UU^\top$ with $U \in \mathbb{R}^{n \times d}$, the problem becomes
\[
\min_{U \in \mathbb{R}^{n \times d}} \|A(UU^\top) - b\|^2.
\]
Under mild assumptions on initialization, the authors show that gradient flow on $U$ converges to a factorization corresponding to the minimal nuclear norm solution of the original problem. This result illustrates how reparameterization fundamentally changes the optimization landscape, leading gradient descent to converge toward more "regularized" solutions.  
Subsequent works have deepened this understanding: \cite{li2018algorithmic} analyze the algorithmic regularization effects arising in such factorized settings; \cite{arora2018optimization} demonstrate that overparameterization can improve conditioning, thereby accelerating convergence; \cite{gunasekar2018characterizing} explore the geometry of the optimization path itself; and \cite{arora2019implicit} extend the analysis to \emph{deep} matrix factorization, linking initialization scale to convergence toward minimal nuclear norm solutions.

\paragraph{Weight normalization} 

Weight normalization, first introduced by \cite{salimans2016weight}, can also be viewed through the lens of optimization dynamics. By decoupling the weight direction and scale, this reparameterization improves the conditioning of the optimization problem and often accelerates convergence. Later analyses, such as \cite{wu2020implicit}, formalize its convergence properties and describe how the induced geometry affects the optimization trajectory. More recently, \cite{chou2024robust} have shown that weight normalization can be calibrated to simultaneously improve robustness and optimization efficiency.

\paragraph{More complex parameterizations}

Beyond simple matrix or vector factorizations, a range of studies has investigated the optimization behavior of more complex architectures. For example, \cite{gunasekar2018implicit} analyze linear convolutional networks, while \cite{allen2019learning,allen2019convergence} establish convergence guarantees for shallow and deep overparameterized networks, respectively. In the case of ReLU networks, \cite{vardi2021implicit} and others have characterized the gradient flow dynamics, highlighting convergence regimes. \cite{marion2024deep} provide convergence analyses for deep linear networks, showing that gradient-based optimization implicitly favors flatter minima. Similarly, \cite{chizat2020implicit} study global convergence of gradient descent in infinitely wide two-layer networks, showing how overparameterization can ensure convergence to near-optimal solutions.

\subsection{Notation}
For any positive integer $L$, the set of integers from $1$ to $L$ is denoted by $[L]$. The coordinate-wise product, also called the Hadamard product, of two vectors $x$ and $y$ in $\mathbb{R}^p$ is denoted $x \odot y$, where for each $i$ in $[p]$, $(x \odot y)_i = x_i y_i$. The vector $x^{\odot L}$ is defined as $\left(x^L_i\right)_{i\in[p]}$ and if $\left(x^j\right)_{j\in[L]}\in\left(\R^p\right)^L$, then $\bigodot_{j=1}^{L}x^j$ refers to $\left(\prod_{j=1}^{L}x^j_i\right)_{i\in[p]}$. We use $|\cdot|$ to denote the absolute value which can be defined for vectors in $\R^p$: for any $x\in\R^p$, $|x|=\left(|x_i|\right)_{i\in[p]}$. Similarly, the sign function denoted $\text{sign}(\cdot)$, the square root function $\sqrt{\cdot}$, the hyperbolic sine $\sinh$, the inverse hyperbolic sine $\text{arcsinh}$ and the logarithm $\log$ are considered as functions from $\R^p$ to $\R^p$ that apply to each component. The Jacobian of $G = \left(G_1, \ldots, G_p\right)$ with respect to $\theta \in \mathbb{R}^k$ is denoted by $J_G(\theta)$ and is defined as:

\[
J_G(\theta) := \left[\frac{\partial G}{\partial \theta_1}; \ldots; \frac{\partial G}{\partial \theta_k}\right] = \left[\begin{array}{ccc}
    \frac{\partial G_1}{\partial \theta_1} & \ldots & \frac{\partial G_1}{\partial \theta_k}\\
    \vdots & \ddots & \vdots \\
    \frac{\partial G_p}{\partial \theta_1} & \ldots & \frac{\partial G_p}{\partial \theta_k}
\end{array}\right].
\]
The Hessian of $G$ is denoted by $\nabla^2 G$. We denote by $\#$ the cardinality of a set, and by $\%$ the modulo operation. The vector $\mathbf{1}\in\R^p$ denotes the unitary vector equal to $\left(1\right)_{i\in[p]}$ and $\lfloor\cdot\rfloor$ represents the floor function. For a vector $v\in\R^p$, $\text{diag}(v)$ denotes the diagonal matrix in $\R^{p\times p}$ having the diagonal equal to $v$.

{
}

\section{Deep Diagonal Linear Networks}\label{sec:3}

\begin{figure}[ht]
\begin{center}
\begin{tikzpicture}[x=2.2cm,y=1cm]
  \foreach \N [count=\lay,remember={\N as \Nprev (initially 0);}]
               in {5,5,5,5,1}{ 
    \foreach \i [evaluate={\y=\N/2-\i; \x=\lay; \prev=int(\lay-1);}]
                 in {1,...,\N}{ 
      \ifnum\lay=1
        \node[mynodeinput] (N\lay-\i) at (\x,\y) {$x^i_\i$};
      \fi
      \ifnum\lay=2
        \node[mynode] (N\lay-\i) at (\x,\y) {$u^1_\i$};
        \draw[thick] (N\prev-\i) -- (N\lay-\i);
      \fi
      \ifnum\lay=3
        \node[mynode] (N\lay-\i) at (\x,\y) {$u^2_\i$};
        \draw[thick] (N\prev-\i) -- (N\lay-\i);
      \fi
      \ifnum\lay=4
        \node[mynode] (N\lay-\i) at (\x+1,\y) {$u^L_\i$};
        \draw[thick,dashed] (N\prev-\i) -- (N\lay-\i);
      \fi
      \ifnum\lay=5
        \node[mynodeoutput] (N\lay-\i) at (\x+1,\y) {$f_\theta(x^i)$};
        \foreach \j in {1,...,\Nprev}{ 
          \draw[thick] (N\prev-\j) -- (N\lay-\i);
        }
      \fi
    }
  }
\end{tikzpicture}
\end{center}
\caption{Representation of a Deep Diagonal Linear Network with $L$ layers in $\R^d$, $d=5$.}
\label{fig:DDLN}
\end{figure}
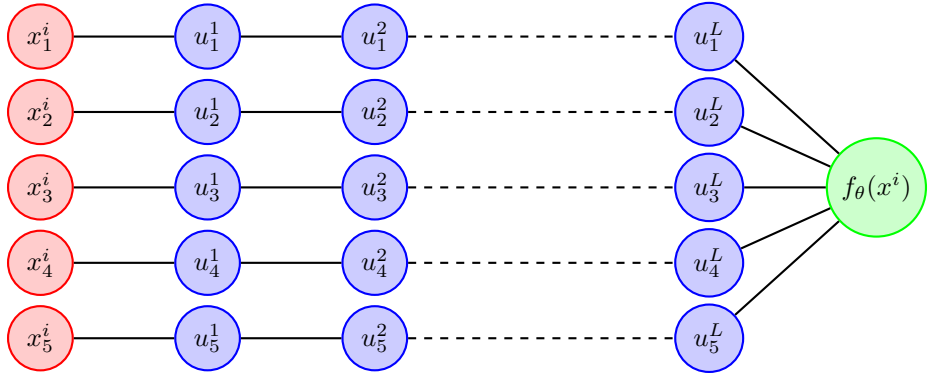

In this section, we state several results on the trajectory of the variable $\theta$ when parameterized by a Deep Diagonal Linear Network and when its parameters are trained by Gradient Flow. More precisely, we overparameterize our system by letting	 
\begin{equation}\label{eq:deep_rep}
\theta=\bigodot_{j=1}^Lu^j,
\end{equation}
where $L\geqslant2$ is the number of layers of the network and each $u^j\in\R^d$ corresponds to the $j$-th layer, as illustrated in Figure \ref{fig:DDLN}. Note that for $L=2$, we recover the classical Diagonal Linear Network. We consider the training of the Deep Diagonal Linear Network by Gradient Flow, namely the minimization of 
\begin{equation*}
    \mathcal{L}\bigg(\bigodot_{j=1}^Lu^j\bigg),
\end{equation*}
which can be written as follows:
\begin{equation}
    \forall j\in[L],~ \forall t\geqslant0,\quad \frac{du^j(t)}{dt}+\nabla_j \mathcal{L}\left(\bigodot_{k=1}^Lu^k(t)\right)=0,
\end{equation}
where  $\left(\nabla_j\mathcal{L}\left(\bigodot_{k=1}^Lu^k\right)\right)_i=\frac{\partial}{\partial u^j_i} \mathcal{L}\left(\bigodot_{k=1}^Lu^k\right)$ for any $j\in[L]$ and $i\in[d]$. By simple computations, we get the following dynamical system:
\begin{equation}\label{eq:GF_Llay}
    \forall j\in[L],~ \forall t\geqslant0,\quad \frac{du^j(t)}{dt}+\left(\bigodot_{k\neq j}u^k(t)\right)\odot\nabla \mathcal{L}(\theta(t))=0.
\end{equation}

Next, we first derive several properties of the trajectories of the weights of the network, and then derive the main result on the dynamic of $\theta$. Convergence guarantees are stated in the last section, under the same assumption on the initialization.

\subsection{On the behavior of the flow}

From \eqref{eq:GF_Llay}, we can prove the following properties on the trajectory of the layers weights.
\begin{proposition}\label{prop:u2}
    Let $\left(u^j\right)_{j\in[L]}$ satisfy \eqref{eq:GF_Llay} and let $\theta=\bigodot_{j=1}^Lu^j$. Then the following statements hold:
    \begin{enumerate}
        \item For any $(j,k)\in[L]\times[L]$, 
    \begin{equation}
        \forall t\geqslant0, ~\frac{d}{dt}\left(u^j\odot u^j\right)(t)=\frac{d}{dt}\left(u^k\odot u^k\right)(t)=-\theta(t)\odot\nabla L(\theta(t)).
    \end{equation}
    In particular,
    \begin{equation}
        \forall t\geqslant0,~u^j(t)^{\odot2}-u^j(0)^{\odot2}=u^k(t)^{\odot2}-u^k(0)^{\odot2}.
    \end{equation}
    \item Let $i\in[d]$. If $j\notin \arg\min_{k\in[L]}|u^k_i(0)|$, then for every  $t>0$ we have that $u_i^j(t) \neq 0$.
    \end{enumerate}
\end{proposition}
\noindent\textbf{Proof.}\\1. This statement is obtained by multiplying \eqref{eq:GF_Llay} by $u^j(t)$ and then by integrating it.\\
2. Let $i\in[d]$ and $j\notin \arg\min_{k\in[L]}|u^k_i(0)|$. We suppose that there exists $T>0$ such that $u_i^j(T)=0$. We define $l\in\arg\min_{k\in[L]}|u^k_i(0)|$ and according to the first claim, for all $t\geqslant0$,
\begin{equation*}
    u^l_i(t)^{2}-u^l_i(0)^{2}=u^j_i(t)^{2}-u^j_i(0)^{2}.
\end{equation*}
In particular, for $t=T$,
\begin{equation*}
    u^l_i(T)^{2}=u^l_i(0)^{2}+u^j_i(T)^{2}-u^j_i(0)^{2}=u^l_i(0)^{2}-u^j_i(0)^{2}<0.
\end{equation*}
Since the above equation cannot be true, we can conclude.\qed\\

The following assumption on the initialization of the network is considered in the following and  useful to  interpret the above result. Assume that, 
\begin{equation}\tag{$\mathcal{A}$}\label{eq:assumption1}
\forall i\in[d],\ ~\#\left(\argmin_{k\in[L]}|u^k_i(0)|\right)=1.\end{equation}
The above condition states that, for each component $i\in[d]$, the node with the minimal absolute value is unique across all layers. For instance, initializing each layer with the same values violates this assumption; while setting them randomly is almost surely valid. This choice of the  initialization  is  weak  compared to other  in the literature, even when considering only two layers. For shallow diagonal linear networks, \cite{woodworth2020kernel} impose a layer to be zero before training while in the context of deep matrix factorization, \cite{arora2019implicit} impose the layers to be balanced i.e.
\begin{equation*}
    \forall (i,j)\in[L]^2, \quad W_i(0)^TW_i(0)=W_j(0)^TW_j(0),
\end{equation*}
where $W_j$ is the matrix containing the weights of the $j$-th layer.

\begin{figure}[ht]
    \centering
    \includegraphics[width=1\linewidth]{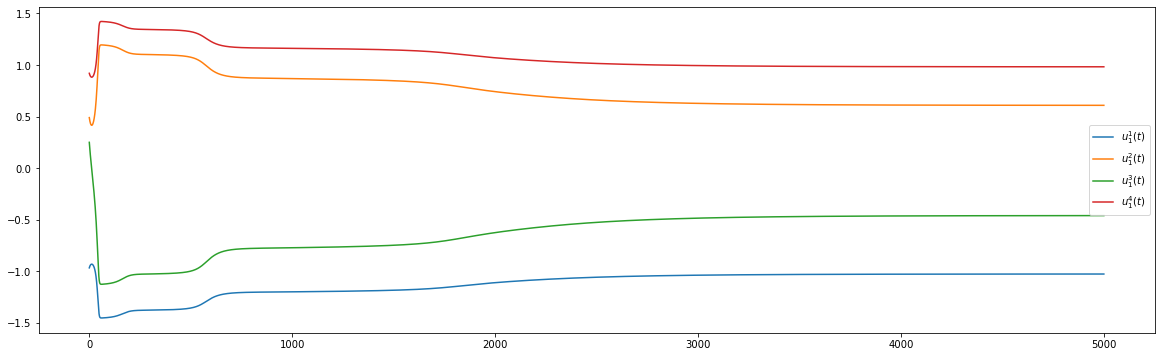}
    \caption{Dynamic of the nodes $(u^j_1(t))_{j\in[L]}$ for a loss function satisfying $\mathcal{L}:\theta\mapsto\|X\theta-y\|^2$ with $X\in\R^{10\times 5}$ and $y\in\R^{10}$ generated randomly, and $L=4$ layers.}
    \label{fig:cross}
\end{figure}

Back to Proposition \ref{prop:u2}, we note that,   by Assumption \ref{eq:assumption1}, the second claim ensures that for each component $i$, only one layer $j$ can cross zero (and this layer is the one with the minimal absolute value at initialization). This phenomenon can be observed in Figure \ref{fig:cross} where for the first component, only the third layer changes its sign throughout the training process. This layer is the one with minimal absolute value at initialization.

We use this observation to rewrite the dynamic of the layers.
\paragraph{Simplifying the system}
Assumption \ref{eq:assumption1}  allows  to define $\left(v^j\right)_{j\in[L]}\in\left(\R^d\right)^L$ as a permutation of $\left(u^j\right)_{j\in[L]}$ in the following way: for any $i\in[d]$,
\begin{equation*}
    v^1_i=u^j_i~\text{where}~j=\arg\min_{k\in[L]}|u^k_i(0)|,
\end{equation*}
and for any $j\in\llbracket2,L\rrbracket$,
\begin{equation*}
    v^j_i=\left\{\begin{aligned}
        &u^1_i~\text{if}~j=\arg\min_{k\in[L]}|u^k_i(0)|.\\
        &u^j_i~\text{else.}
    \end{aligned}\right.
\end{equation*}
In other words, $\left(v^j\right)_{j\in[L]}$ is a permutation of $\left(u^j\right)_{j\in[L]}$ where the first layer gathers all minimal absolute values at initialization. Note that we have $\theta=\bigodot_{j=1}^Lv^j$ and it is trivial to show that the  properties discussed above for  $\left(u^j\right)_{j\in[L]}$  still hold for $\left(v^j\right)_{j\in[L]}$.\\

Since we assume that \ref{eq:assumption1} holds, the second claim of Proposition \ref{prop:u2} ensures that for any $j\in\llbracket2,L\rrbracket$,
\begin{equation*}
    \forall t\geqslant0,~\text{sign}\left(v^j(t)\right)=\text{sign}\left(v^j(0)\right),
\end{equation*}
where $\text{sign}:\R^d\rightarrow\R^d$ returns the sign for every component. By  the first claim of Proposition \ref{prop:u2}, we get that
\begin{equation*}
    \forall j\in\llbracket2,L\rrbracket,~\forall t\geqslant0,~v^j(t)=\text{sign}\left(v^j(0)\right)\odot\sqrt{v^1(t)^{\odot2}+v^j(0)^{\odot2}-v^1(0)^{\odot2}}.
\end{equation*}
By rewriting \eqref{eq:GF_Llay} for the first layer, we have that:
\begin{equation}\label{eq:v1}
    \forall t\geqslant0,~\frac{dv^1(t)}{dt}+\text{sign}\left(\bigodot_{j\neq 1}v^j(0)\right)\odot\left(\bigodot_{j\neq1}\sqrt{v^1(t)^{\odot2}+\Delta_j}\right)\odot\nabla L(\theta(t))=0,
\end{equation}
where $\Delta_j=v^j(0)^{\odot2}-v^1(0)^{\odot2}$ for any $j\in[d]$. Since $$\theta(t)=\text{sign}\left(\bigodot_{j= 1}^Lv^j(0)\right)\odot\left(\bigodot_{j=1}^L\sqrt{v^1(t)^{\odot2}+\Delta_j}\right),$$ we can see that {the flow of $\theta$ is entirely determined by that of $v^1$ and the initialization of the network}.

\paragraph{Deducing a Mirror Flow?}  The above observations could suggest to consider the strategy proposed by \cite{woodworth2020kernel} to study shallow Diagonal Linear Networks. There,  a suitable expression for $\theta$ is leveraged to  derive an implicit bias  linked to a Mirror Flow (for further details, see Appendix \ref{sec:DLN}; for another example to which it applies, see Appendix \ref{app:red}). However, if $L>2$, \eqref{eq:v1} does not allow us to get an analytical expression of $v^1(t)$ and thus of $\theta(t)$. As a consequence, this approach cannot be applied in this situation. Note that in the case $L=2$, \eqref{eq:v1} gives for any $t\geqslant0$,
\begin{equation}
    \frac{dv^1(t)}{dt}+\text{sign}\left(v^2(0)\right)\odot\sqrt{v^1(t)^{\odot2}+\Delta}\odot\nabla L(\theta(t))=0,
\end{equation}
with $\Delta = v^2(0)^{\odot2}-v^1(0)^{\odot2}$, allowing to compute $\theta$ and recover the analysis of \cite{woodworth2020kernel}.

\subsection{Mirror Flow for Deep Diagonal Linear Network}\label{sec:thm1}

Since  our setting does not fit the framework of \cite{woodworth2020kernel}, we base our analysis on the approach proposed in\cite{li2022implicit} to prove the following theorem.

\begin{theorem}\label{thm:1}
    Let $\theta\in\R^d$ be parameterized as in equartion~\eqref{eq:deep_rep} with  $\left(u^j\right)_{j\in[L]}$ following the dynamic  in \eqref{eq:GF_Llay}. If the initialization of the network satisfies \ref{eq:assumption1}, then $\theta$ follows a Mirror Flow dynamic, i.e. there exists a convex Legendre function $\mathcal{Q}:\R^d\rightarrow\R$ such that:
    \begin{equation*}
        \forall t\geqslant0,\quad\frac{d\nabla \mathcal{Q}(\theta(t))}{dt}+\nabla \mathcal{L}(\theta(t))=0.
    \end{equation*}
    {In particular,
    \begin{equation}\label{eq:mirror_rate}
    \mathcal{L}(\theta(t))-\mathcal{L}^* \leq \frac{D_Q(\theta^*,\theta_0)}{t}, 
    \end{equation}
    where $\theta^*\in\argmin_{\theta\in\mathbb{R}^d} \mathcal{L}(\theta)$ and $D_Q(\theta^*,\theta)=Q(\theta^*)-Q(\theta)-\langle \nabla Q(\theta),\theta^*-\theta\rangle$.
}\end{theorem}

\noindent\textbf{Proof.}
The proof of this claim consists in showing that the reparameterization induced by Deep Diagonal Linear Networks satisfies the hypotheses necessary to the application of \cite[Theorem~4.8]{li2022implicit}. It is first necessary to introduce some notation.

Let $\theta\in\R^d$ be parameterized as $\theta=\bigodot_{j=1}^Lu^j$ where $\left(u^j\right)_{j\in[L]}$. We denote $\vu\in\R^{L\times d}$ the entire set of parameters of the network. More specifically, we have
\begin{equation*}
    \forall n\in[Ld],~\vu_i=u^{(i-1)\%L+1}_{\lfloor (i-1)/L\rfloor+1},
\end{equation*}
meaning that we can write
\begin{equation*}
    \vu=\left(u^1_1~u^2_1~\dots~u^L_1~u^1_2~\dots\dots\dots~u^1_d~\dots~u^L_d\right)^T.
\end{equation*}
It is then possible to write $\theta$ according to $\vu$ via the parameterization function $G:\R^{L\times d}\rightarrow\R^d$ defined as follows
\begin{equation}
    \forall w\in\R^{L\times d},~\forall i\in[d],~G_i(w)=\prod_{j=(i-1)L+1}^{iL}w_j.
\end{equation}
Showing the desired claim then requires to prove that $G:\mathcal{M}\rightarrow\R^d$ is a commuting and regular parameterization for a well-chosen smooth manifold $\mathcal{M}$ of $\R^{L\times d}$. Then, under an additional technical assumption that we discuss and if $\vu(0)\in\mathcal{M}$, \cite[Theorem~4.8]{li2022implicit} ensures that the trajectory of $\theta$ is the solution of a Mirror Flow. The detailed computations are given in Appendix \ref{app:thm1}.

\qed\\ 
\ \\

The above results shows that, under mild assumptions on the inizialization,   Gradient Flow on the reparametrization $(u^j)_{j\in [L]}$  generates a trajectory in the effective parameters $\theta(t)=\odot_{j=1}^Lu^j(t)$ that is the solution of a Mirror Flow on the function $\mathcal{L}$ with entropy $\mathcal{Q}$. Our approach ensure that $\mathcal{Q}$ is convex and so, for instance, that the quantity $\mathcal{L}(\theta(t))$ is non-increasing along time. Indeed, in the following quantitative  convergence guarantees will also be derived.  However,  our proof strategy does not lead to an explicit expression for the entropy $\mathcal{Q}$ (and hence  for the implicit bias induced by the reparametrization). Interestingly,  the results in \cite{yununifying} allow to derive the expression of the  entropy $\mathcal{Q}$ for a particular   initialization, namely when $u^L(0) = 0$ and $u^j(0) = \alpha \bar{u}$ for every $j \in [L-1]$, with $\alpha > 0$ and $\bar{u} \in \mathbb{R}^d$ satisfying $\bar{u}_i \neq 0$ for all $i \in [d]$.  Indeed,  under this initialization,  and when $\mathcal{L}$ is the least squares loss, the trajectory $\theta(t)$ converges to $\theta_\infty$, a global minimizer. Moreover, the limiting solution minimizes a distance that behaves like the $\ell_1$-norm as $\alpha^L \to 0$, and like the $\ell_2$-norm as $\alpha^L \to \infty$. For further details, we refer the interested reader to \cite[Corollary~5]{yununifying}. It remains unclear whether the  formulation introduced in this paper can be extended to accommodate the choice of initialization as in  assumption $\mathcal{A}$. This question is left as a direction for future research. Instead, we next derive suitable convergence guarantees.

\subsection{Convergence guarantees}

We derive additional convergence results on the trajectory of $\theta$ under  assumption \ref{eq:assumption1}. First, note that $\theta$ is a solution of the following dynamical system
\begin{equation}\label{eq:dyn_theta_Ll}
    \forall t\geqslant 0,\quad\dot{\theta}(t)+\sum_{j=1}^L\left(\bigodot_{k\neq j}u^k(t)^{\odot2}\right)\odot\nabla \mathcal{L}(\theta(t))=0.
\end{equation}
We can easily write this equation in the following way:
\begin{equation}\label{eq:dyn_theta_Ll2}
    \forall t\geqslant 0,\quad\dot{\theta}(t)+M(t)\nabla \mathcal{L}(\theta(t))=0,
\end{equation}
where $M:t\mapsto \text{diag}\left(\left(\sum_{j=1}^L\left(\prod_{k\neq j}u^k_i(t)^{2}\right)\right)_{i\in[d]}\right)$.\\
Since we  assumption \ref{eq:assumption1} holds, the second claim of Proposition \ref{prop:u2} ensures that for any $i\in[d]$, if $j\notin\arg\min_{k\in[L]}|u^k_i(0)|$, then $|u^j_i(t)|>0$ for all $t\geqslant0$. More precisely, using the first claim of Proposition \ref{prop:u2}, we obtain that $u^j_i(t)^{2}\geqslant u^j_i(0)^{2}-u^l_i(0)^{2}$ where $l=\arg\min_{k\in[L]}|u^k_i(0)|$. Therefore, for any $i\in[d]$,
\begin{equation}
    \begin{aligned}\forall t\geqslant0,~\sum_{j=1}^L\left(\prod_{k\neq j}u^k_i(t)^{2}\right)&=\sum_{j=1,~j\neq l}^L\left(\prod_{k\neq j}u^k_i(t)^{2}\right)+\prod_{k\neq l}u^k_i(t)^{2}\\
    &\geqslant 0+\prod_{k\neq l}( u^k_i(0)^{2}-u^l_i(0)^{2})>0.
    \end{aligned}
\end{equation}
Then, we get the following properties:
\begin{itemize}
    \item $M(t)$ is invertible for all $t\geqslant0$ and 
    $$M^{-1}(t)=\text{diag}\left(\left(\left(\sum_{j=1}^L\left(\prod_{k\neq j}u^k_i(t)^{2}\right)\right)^{-1}\right)_{i\in[d]}\right).$$
    In particular, 
    \begin{equation*}
        \forall t\geqslant0, \quad M^{-1}(t)\dot{\theta}(t)+\nabla \mathcal{L}(t)=0,
    \end{equation*}
    which means that the mirror map $\mathcal{Q}$ from Theorem \ref{thm:1} satisfies for any $t\geqslant 0$
    \begin{equation*}
        \frac{d\nabla\mathcal{Q}(\theta(t))}{dt}=M^{-1}(t)\dot{\theta}(t).
    \end{equation*}
    \item The smallest eigenvalue $\lambda_{min}(M(t))$ of $M(t)$ admits a lower bound that does not depend on $t$:
    \begin{equation}\label{eq:lambda_min}
        \lambda_{min}(M(t))\geqslant \min_{i\in[d]}\prod_{k\neq k_i}( u^k_i(0)^{2}-u^{k_i}_i(0)^{2}),
    \end{equation}
    where $k_i=\arg\min_{k\in[L]}|u^k_i(0)|$ for any $i\in[d]$.
\end{itemize}

Then, we get the following result which applies, for instance, if $\mathcal{L}$ is the squared loss function.

\begin{theorem}
    Let $\theta\in\R^d$ be parameterized as $\theta=\bigodot_{j=1}^Lu^j$ and $\left(u^j\right)_{j\in[L]}$ follow the dynamic described in \eqref{eq:GF_Llay}. Suppose that $\arg\min_{\theta\in\R^d}\mathcal{L}(\theta)\neq\emptyset$ and let $\mathcal{L}^*$ denote $\min_{\theta\in\R^d}\mathcal{L}(\theta)$. If the initialization of the network satisfies \ref{eq:assumption1} and the loss function $\mathcal{L}$ satisfies a Polyak-\L{}ojasiewicz property with parameter $\mu>0$, then for all $t\geqslant0$,
    \begin{equation*}
        \mathcal{L}(\theta(t))-\mathcal{L}^*\leqslant e^{-2\sigma\mu t}\left(\mathcal{L}(\theta(0))-\mathcal{L}^*\right),
    \end{equation*}
    where $\sigma=\min_{i\in[d]}\prod_{k\neq k_i}( u^k_i(0)^{2}-u^{k_i}_i(0)^{2})$ and $k_i=\arg\min_{k\in[L]}|u^k_i(0)|$ for any $i\in[d]$.
\end{theorem}
\noindent\textbf{Proof.}\\
For all $t\geqslant0$, we have that:
\begin{equation*}
    \begin{aligned}
        \frac{d}{dt}\left(\mathcal{L}(\theta(t))-\mathcal{L}^*\right)&=\left\langle\nabla \mathcal{L}(\theta(t)),\dot{\theta}(t)\right\rangle\\
        &=-\left\langle\nabla \mathcal{L}(\theta(t)),M(t)\nabla \mathcal{L}(\theta(t))\right\rangle\\
        &\leqslant -\lambda_{min}(M(t))\|\nabla \mathcal{L}(\theta(t))\|^2\\
        &\leqslant -\sigma\|\nabla \mathcal{L}(\theta(t))\|^2,
    \end{aligned}
\end{equation*}
where we use the lower bound $\sigma=\min_{i\in[d]}\prod_{k\neq k_i}( u^k_i(0)^{2}-u^{k_i}_i(0)^{2})$ obtained in \eqref{eq:lambda_min}. Supposing that $\mathcal{L}$ satisfies the Polyak-\L{}ojasiewicz inequality, there exists $\mu>0$ such that:
\begin{equation*}
    \forall \theta\in\R^d,\quad2\mu\left(\mathcal{L}(\theta)-\mathcal{L}^*\right)\leqslant \|\nabla \mathcal{L}(\theta)\|^2.
\end{equation*}
In particular, this guarantees that:
\begin{equation*}
    \frac{d}{dt}\left(\mathcal{L}(\theta(t))-\mathcal{L}^*\right)\leqslant -2\sigma\mu\left(\mathcal{L}(\theta(t))-\mathcal{L}^*\right),
\end{equation*}
and as a consequence:
\begin{equation}
    \forall t\geqslant0,\quad \mathcal{L}(\theta(t))-\mathcal{L}^*\leqslant e^{-2\sigma\mu t} \left(\mathcal{L}(\theta(0))-\mathcal{L}^*\right).
\end{equation}\qed\\
The above bound on the suboptimality gap reveals that the initialization of the network significantly affects the convergence rate of its training and that $\sigma=\min_{i\in[d]}\prod_{k\neq k_i}( u^k_i(0)^{2}-u^{k_i}_i(0)^{2})$ plays an important role.  Consider for example the case where $\mathcal{L}$ is $\mu$ strongly convex and hence satisfies the $\mu$ PL inequality. In this setting, vanilla gradient flow  applied to the effective parameters gives a rate $e^{-2\mu t}(\mathcal{L}(\theta_0)-\mathcal{L}^*)$. Instead, considering the overparameterization~\eqref{eq:deep_rep} we obtain in the new variables a problem which is non convex in general. Despite this, the previous result shows that  gradient flow on such non convex problem may result in the original variable $\theta$ in a faster converegence rate,  as long as the  initialization  is such that  $\sigma>1$.


This becomes particularly clear when considering  the initialization  considered in \cite{yununifying}.
\begin{corollary}
    Let $\theta\in\R^d$ be parameterized as in equation~\eqref{eq:deep_rep} and $\left(u^j\right)_{j\in[L]}$ follow the dynamic  in \eqref{eq:GF_Llay}. Suppose that $\arg\min_{\theta\in\R^d}\mathcal{L}(\theta)\neq\emptyset$ and let $\mathcal{L}^*$ denote $\min_{\theta\in\R^d}\mathcal{L}(\theta)$. If the initialization is such that $u^L(0) = 0$ and $u^j(0) = \alpha \bar{u}$ for every $j \in [L-1]$, with $\alpha > 0$ and $\bar{u} \in \mathbb{R}^d$ satisfying $|\bar{u}_i|^2 \geqslant\lambda> 0$ for all $i \in [d]$, and the loss function $\mathcal{L}$ satisfies a Polyak-\L{}ojasiewicz property with parameter $\mu>0$, then for all $t\geqslant0$,
    \begin{equation*}
        \mathcal{L}(\theta(t))-\mathcal{L}^*\leqslant e^{-2\sigma\mu t}\left(\mathcal{L}(\theta(0))-\mathcal{L}^*\right), \qquad \sigma=(\alpha^{2}\lambda)^{L-1}.
    \end{equation*}
\end{corollary}
\noindent\textbf{Proof.}\\
The initialization of the network directly ensures that assumption \ref{eq:assumption1} is satisfied and the explicit computation of $\sigma$ is straightforward.\qed\\

\begin{figure}[ht]
    \centering
    \includegraphics[width=1\linewidth]{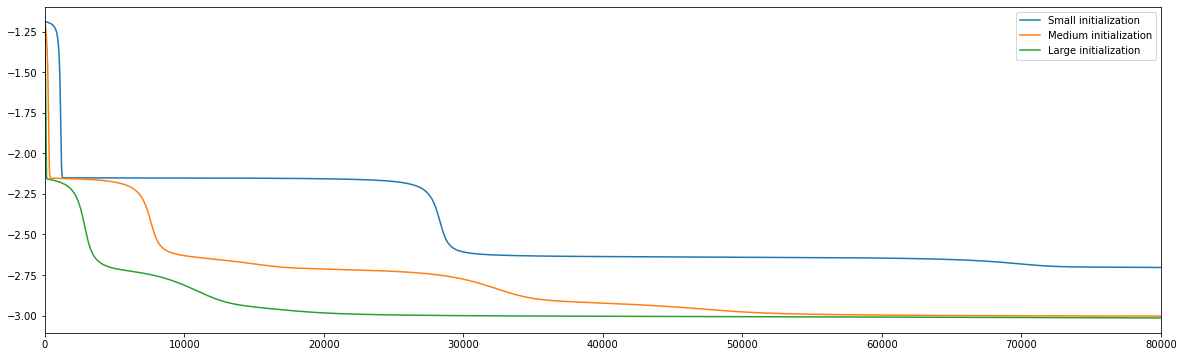}
    \caption{Evolution of $\log\left(\mathcal{L}(\theta(t))-\mathcal{L}^*\right)$ according to time for three $6$-layer networks with different initialization. The loss function is defined as $\mathcal{L}:\theta\mapsto\|X\theta-y\|^2$ with $X\in\R^{10\times 8}$ and $y\in\R^{10}$ generated randomly. Each network is initialized with a first layer having components equal to $0$. The initial value of the remaining layers of the first network (in blue) is generated randomly, while that of the second (in orange) and the third (in green) are respectively equal to $1.4$ and $1.8$ times the values of the first component-wise.}
    \label{fig:loss}
\end{figure}

In this case, if $\alpha^2\lambda>1$ then  $\sigma$ is increasing as a function of the number of layers, and this emphasizes the beneficial effect of deep overparametrizations on the convergence rate.   Figure \ref{fig:loss} hihglights the impact of the initial amplitude of the weights, showing that larger values at initialization lead to a better convergence rate.

The above results highlights a  phenomenon documented in the literature for similar models (e.g. see \cite{chou2023more,chou2024robust}) and may indicate that initializing the network with large values would be beneficial. However, it is also well known that the initialization of a network could be chosen small enough to induce sparsity of the approximated solutions, see \cite{gunasekar2017implicit, arora2019implicit,woodworth2020kernel,moroshko2020implicit,NEURIPS2019_ae614c55}. Indeed, to address the challenge of slow training, \cite{chou2024robust} suggest employing weight normalization, a reparameterization technique that promotes comparable sparsity in the solution while ensuring reasonable convergence rates.

\section{Conclusion and Perspectives}

In this work, we analyzed the optimization dynamics of Deep Diagonal Linear Networks by studying the trajectory of the effective parameter induced by gradient flow on the layers weights variables. This approach allows  to identify mild initialization conditions under which the dynamics in the effective parameter space follow a mirror-flow structure and to derive convergence guarantees, including linear convergence under a Polyak–Łojasiewicz condition. Our analysis  highlights the detrimental effect of excessively small initializations on the convergence speed.
Although the model we consider corresponds to a simple learning architecture, its dynamics and analysis remain surprisingly intricate. An open question is whether the entropy function associated with the induced mirror flow can be characterized explicitly. This  would further clarify the underlying implicit geometry as well as the induced bias. Clearly considering non linear models, e.g. homogenous nonlinearities,  would be a step towards more practical learning architectures
Finally, we note that   from an optimization standpoint, incorporating stochasticity, momentum, or alternative update rules offers additional avenues for understanding how algorithmic choices interact with overparameterized parametrizations.

\section*{Acknowledgements}

H. L., L. R., C. M. and S. V. acknowledge the financial support of the European Research Council (grant SLING 819789), the European Commission (ELIAS 101120237), the US Air Force Office of Scientific Research (FA8655-22-1-7034), the Ministry of Education, University and Research (grant ML4IP R205T7J2KP; the European Commission (grant TraDE-OPT 861137); grant BAC FAIR PE00000013 funded by the EU - NGEU) and the Center for Brains, Minds and Machines (CBMM), funded by NSF STC award CCF-1231216. The research by C. M. and S. V. has been supported by the MUR Excellence Department Project awarded to Dipartimento di Matematica, Universita di Genova, CUP D33C23001110001. C. M. and S. V. are members of the Gruppo Nazionale per l’Analisi Matematica, la Probabilità e le loro Applicazioni (GNAMPA) of the Istituto Nazionale di Alta Matematica (INdAM).  This work represents only the view of the authors. The European Commission  and the other organizations are not responsible for any use that may be made of the information it contains.

\bibliography{ref.bib}

@inproceedings{woodworth2020kernel,
  title={Kernel and rich regimes in overparametrized models},
  author={Woodworth, Blake and Gunasekar, Suriya and Lee, Jason D and Moroshko, Edward and Savarese, Pedro and Golan, Itay and Soudry, Daniel and Srebro, Nathan},
  booktitle={Conference on Learning Theory},
  pages={3635--3673},
  year={2020},
  organization={PMLR}
}

@inproceedings{papazov2024leveraging,
  title={Leveraging Continuous Time to Understand Momentum When Training Diagonal Linear Networks},
  author={Papazov, Hristo and Pesme, Scott and Flammarion, Nicolas},
  booktitle={International Conference on Artificial Intelligence and Statistics},
  pages={3556--3564},
  year={2024},
  organization={PMLR}
}

@article{li2022implicit,
  title={Implicit bias of gradient descent on reparametrized models: On equivalence to mirror descent},
  author={Li, Zhiyuan and Wang, Tianhao and Lee, Jason D and Arora, Sanjeev},
  journal={Advances in Neural Information Processing Systems},
  volume={35},
  pages={34626--34640},
  year={2022}
}

@article{marion2024deep,
  title={Deep linear networks for regression are implicitly regularized towards flat minima},
  author={Marion, Pierre and Chizat, L{\'e}na{\"\i}c},
  journal={arXiv preprint arXiv:2405.13456},
  year={2024}
}

@article{gunasekar2017implicit,
  title={Implicit regularization in matrix factorization},
  author={Gunasekar, Suriya and Woodworth, Blake E and Bhojanapalli, Srinadh and Neyshabur, Behnam and Srebro, Nati},
  journal={Advances in neural information processing systems},
  volume={30},
  year={2017}
}

@article{chou2023induce,
  title={How to induce regularization in generalized linear models: A guide to reparametrizing gradient flow},
  author={Chou, Hung-Hsu and Maly, Johannes and St{\"o}ger, Dominik},
  journal={arXiv preprint arXiv:2308.04921},
  year={2023}
}

@article{hoff2017lasso,
  title={Lasso, fractional norm and structured sparse estimation using a Hadamard product parametrization},
  author={Hoff, Peter D},
  journal={Computational Statistics \& Data Analysis},
  volume={115},
  pages={186--198},
  year={2017},
  publisher={Elsevier}
}

@article{poon2023smooth,
  title={Smooth over-parameterized solvers for non-smooth structured optimization},
  author={Poon, Clarice and Peyr{\'e}, Gabriel},
  journal={Mathematical programming},
  volume={201},
  number={1},
  pages={897--952},
  year={2023},
  publisher={Springer}
}

@article{zhao2022high,
  title={High-dimensional linear regression via implicit regularization},
  author={Zhao, Peng and Yang, Yun and He, Qiao-Chu},
  journal={Biometrika},
  volume={109},
  number={4},
  pages={1033--1046},
  year={2022},
  publisher={Oxford University Press}
}

@article{vaskevicius2019implicit,
  title={Implicit regularization for optimal sparse recovery},
  author={Vaskevicius, Tomas and Kanade, Varun and Rebeschini, Patrick},
  journal={Advances in Neural Information Processing Systems},
  volume={32},
  year={2019}
}

@article{chou2023more,
  title={More is less: inducing sparsity via overparameterization},
  author={Chou, Hung-Hsu and Maly, Johannes and Rauhut, Holger},
  journal={Information and Inference: A Journal of the IMA},
  volume={12},
  number={3},
  pages={1437--1460},
  year={2023},
  publisher={Oxford University Press}
}

@article{chou2022non,
  title={Non-negative least squares via overparametrization},
  author={Chou, Hung-Hsu and Maly, Johannes and Verdun, Claudio Mayrink},
  journal={arXiv preprint arXiv:2207.08437},
  year={2022}
}

@article{chou2024robust,
  title={Robust implicit regularization via weight normalization},
  author={Chou, Hung-Hsu and Rauhut, Holger and Ward, Rachel},
  journal={Information and Inference: A Journal of the IMA},
  volume={13},
  number={3},
  pages={iaae022},
  year={2024},
  publisher={Oxford University Press}
}

@inproceedings{amid2020winnowing,
  title={Winnowing with gradient descent},
  author={Amid, Ehsan and Warmuth, Manfred K},
  booktitle={Conference on Learning Theory},
  pages={163--182},
  year={2020},
  organization={PMLR}
}

@article{vaskevicius2020statistical,
  title={The statistical complexity of early-stopped mirror descent},
  author={Vaskevicius, Tomas and Kanade, Varun and Rebeschini, Patrick},
  journal={Advances in Neural Information Processing Systems},
  volume={33},
  pages={253--264},
  year={2020}
}

@article{alvarez2004hessian,
  title={Hessian Riemannian gradient flows in convex programming},
  author={Alvarez, Felipe and Bolte, J{\'e}r{\^o}me and Brahic, Olivier},
  journal={SIAM journal on control and optimization},
  volume={43},
  number={2},
  pages={477--501},
  year={2004},
  publisher={SIAM}
}

@inproceedings{azulay2021implicit,
  title={On the implicit bias of initialization shape: Beyond infinitesimal mirror descent},
  author={Azulay, Shahar and Moroshko, Edward and Nacson, Mor Shpigel and Woodworth, Blake E and Srebro, Nathan and Globerson, Amir and Soudry, Daniel},
  booktitle={International Conference on Machine Learning},
  pages={468--477},
  year={2021},
  organization={PMLR}
}

@article{pesme2021implicit,
  title={Implicit bias of sgd for diagonal linear networks: a provable benefit of stochasticity},
  author={Pesme, Scott and Pillaud-Vivien, Loucas and Flammarion, Nicolas},
  journal={Advances in Neural Information Processing Systems},
  volume={34},
  pages={29218--29230},
  year={2021}
}

@inproceedings{gunasekar2018characterizing,
  title={Characterizing implicit bias in terms of optimization geometry},
  author={Gunasekar, Suriya and Lee, Jason and Soudry, Daniel and Srebro, Nathan},
  booktitle={International Conference on Machine Learning},
  pages={1832--1841},
  year={2018},
  organization={PMLR}
}

@inproceedings{nacson2022implicit,
  title={Implicit bias of the step size in linear diagonal neural networks},
  author={Nacson, Mor Shpigel and Ravichandran, Kavya and Srebro, Nathan and Soudry, Daniel},
  booktitle={International Conference on Machine Learning},
  pages={16270--16295},
  year={2022},
  organization={PMLR}
}

@inproceedings{vardi2021implicit,
  title={Implicit regularization in relu networks with the square loss},
  author={Vardi, Gal and Shamir, Ohad},
  booktitle={Conference on Learning Theory},
  pages={4224--4258},
  year={2021},
  organization={PMLR}
}

@article{salimans2016weight,
  title={Weight normalization: A simple reparameterization to accelerate training of deep neural networks},
  author={Salimans, Tim and Kingma, Durk P},
  journal={Advances in neural information processing systems},
  volume={29},
  year={2016}
}

@article{wu2020implicit,
  title={Implicit regularization and convergence for weight normalization},
  author={Wu, Xiaoxia and Dobriban, Edgar and Ren, Tongzheng and Wu, Shanshan and Li, Zhiyuan and Gunasekar, Suriya and Ward, Rachel and Liu, Qiang},
  journal={Advances in Neural Information Processing Systems},
  volume={33},
  pages={2835--2847},
  year={2020}
}

@article{moroshko2020implicit,
  title={Implicit bias in deep linear classification: Initialization scale vs training accuracy},
  author={Moroshko, Edward and Woodworth, Blake E and Gunasekar, Suriya and Lee, Jason D and Srebro, Nati and Soudry, Daniel},
  journal={Advances in neural information processing systems},
  volume={33},
  pages={22182--22193},
  year={2020}
}

@article{allen2019learning,
  title={Learning and generalization in overparameterized neural networks, going beyond two layers},
  author={Allen-Zhu, Zeyuan and Li, Yuanzhi and Liang, Yingyu},
  journal={Advances in neural information processing systems},
  volume={32},
  year={2019}
}

@inproceedings{allen2019convergence,
  title={A convergence theory for deep learning via over-parameterization},
  author={Allen-Zhu, Zeyuan and Li, Yuanzhi and Song, Zhao},
  booktitle={International conference on machine learning},
  pages={242--252},
  year={2019},
  organization={PMLR}
}

@inproceedings{chizat2020implicit,
  title={Implicit bias of gradient descent for wide two-layer neural networks trained with the logistic loss},
  author={Chizat, Lenaic and Bach, Francis},
  booktitle={Conference on learning theory},
  pages={1305--1338},
  year={2020},
  organization={PMLR}
}

@article{gunasekar2018implicit,
  title={Implicit bias of gradient descent on linear convolutional networks},
  author={Gunasekar, Suriya and Lee, Jason D and Soudry, Daniel and Srebro, Nati},
  journal={Advances in neural information processing systems},
  volume={31},
  year={2018}
}

@inproceedings{li2018algorithmic,
  title={Algorithmic regularization in over-parameterized matrix sensing and neural networks with quadratic activations},
  author={Li, Yuanzhi and Ma, Tengyu and Zhang, Hongyang},
  booktitle={Conference On Learning Theory},
  pages={2--47},
  year={2018},
  organization={PMLR}
}

@article{arora2019implicit,
  title={Implicit regularization in deep matrix factorization},
  author={Arora, Sanjeev and Cohen, Nadav and Hu, Wei and Luo, Yuping},
  journal={Advances in Neural Information Processing Systems},
  volume={32},
  year={2019}
}

@inproceedings{arora2018optimization,
  title={On the optimization of deep networks: Implicit acceleration by overparameterization},
  author={Arora, Sanjeev and Cohen, Nadav and Hazan, Elad},
  booktitle={International conference on machine learning},
  pages={244--253},
  year={2018},
  organization={PMLR}
}

@misc{ouyang2024kurdykalojasiewiczexponenthadamardparametrization,
      title={Kurdyka-{\L}ojasiewicz exponent via Hadamard parametrization}, 
      author={Wenqing Ouyang and Yuncheng Liu and Ting Kei Pong and Hao Wang},
      year={2024},
      eprint={2402.00377},
      archivePrefix={arXiv},
      primaryClass={math.OC},
      url={https://arxiv.org/abs/2402.00377}, 
}

@inproceedings{NEURIPS2019_ae614c55,
 author = {Chizat, L\'{e}na\"{\i}c and Oyallon, Edouard and Bach, Francis},
 booktitle = {Advances in Neural Information Processing Systems},
 editor = {H. Wallach and H. Larochelle and A. Beygelzimer and F. d\textquotesingle Alch\'{e}-Buc and E. Fox and R. Garnett},
 pages = {},
 publisher = {Curran Associates, Inc.},
 title = {On Lazy Training in Differentiable Programming},
 url = {https://proceedings.neurips.cc/paper_files/paper/2019/file/ae614c557843b1df326cb29c57225459-Paper.pdf},
 volume = {32},
 year = {2019}
}

@inproceedings{mathieu,
 author = {Even, Mathieu and Pesme, Scott and Gunasekar, Suriya and Flammarion, Nicolas},
 booktitle = {Advances in Neural Information Processing Systems},
 editor = {A. Oh and T. Naumann and A. Globerson and K. Saenko and M. Hardt and S. Levine},
 pages = {29406--29448},
 publisher = {Curran Associates, Inc.},
 title = {(S)GD over Diagonal Linear Networks: Implicit bias, Large Stepsizes and Edge of Stability},
 url = {https://proceedings.neurips.cc/paper_files/paper/2023/file/5da6ce80e97671b70c01a2e703b868b3-Paper-Conference.pdf},
 volume = {36},
 year = {2023}
}

@inproceedings{yununifying,
  title={A unifying view on implicit bias in training linear neural networks},
  author={Yun, Chulhee and Krishnan, Shankar and Mobahi, Hossein},
  booktitle={International Conference on Learning Representations},
  year = {2021}
}

@article{marcotte2023abide,
  title={Abide by the law and follow the flow: Conservation laws for gradient flows},
  author={Marcotte, Sibylle and Gribonval, R{\'e}mi and Peyr{\'e}, Gabriel},
  journal={Advances in neural information processing systems},
  volume={36},
  pages={63210--63221},
  year={2023}
}

@article{marcotte2025transformative,
  title={Transformative or Conservative? Conservation laws for ResNets and Transformers},
  author={Marcotte, Sibylle and Gribonval, R{\'e}mi and Peyr{\'e}, Gabriel},
  journal={arXiv preprint arXiv:2506.06194},
  year={2025}
}

@article{marcotte2024keep,
  title={Keep the momentum: Conservation laws beyond euclidean gradient flows},
  author={Marcotte, Sibylle and Gribonval, R{\'e}mi and Peyr{\'e}, Gabriel},
  journal={arXiv preprint arXiv:2405.12888},
  year={2024}
}

@article{saxe2013exact,
  title={Exact solutions to the nonlinear dynamics of learning in deep linear neural networks},
  author={Saxe, Andrew M and McClelland, James L and Ganguli, Surya},
  journal={arXiv preprint arXiv:1312.6120},
  year={2013}
}

@article{du2018algorithmic,
  title={Algorithmic regularization in learning deep homogeneous models: Layers are automatically balanced},
  author={Du, Simon S and Hu, Wei and Lee, Jason D},
  journal={Advances in neural information processing systems},
  volume={31},
  year={2018}
}
\bibliographystyle{apalike}

\appendix
\section{Appendix}

\subsection{Background: A direct strategy to analyze Diagonal Linear Networks}\label{sec:DLN}

In this section, we briefly state the strategy proposed by \cite{woodworth2020kernel} to analyze the training of a Diagonal Linear Network through Gradient Flow. The proposed approach to derive an implicit bias benefits from its simplicity and is general enough to be extended to other simple models. To emphasize this point, we apply it also to Deep Redundant Linear Network in Appendix \ref{app:red}.

We are interested in solving the problem:
\begin{equation*}
    \min_{\theta\in\R^d}\mathcal{L}(\theta),
\end{equation*}
for some loss function $\mathcal{L}:\R^{d}\rightarrow\R$, where the loss is minimal if $X\theta=y$, $X:\R^d\rightarrow\R^n$ and $y\in\R^n$ denoting some data input and output. We look at this problem through the lens of reparameterization, i.e. we consider the change of variable $\theta=u\odot v$ which corresponds to a Diagonal Linear Network. Such a network is then trained using a Gradient Flow on the new parameters: 
\begin{equation}
    \left\{\begin{aligned}
        &\dot{u}(t)+v(t)\odot\nabla\mathcal{L}(\theta(t))=0\\
        &\dot{v}(t)+u(t)\odot\nabla\mathcal{L}(\theta(t))=0.
    \end{aligned}\right.
\end{equation}
Further computations, detailed in Appendix \ref{app:DLN}, show that the variable $\theta$ can be written in the following way:
\begin{equation}\label{eq:Psi2}
    \theta(t)=\Psi(\xi(t)):=\frac{\left|u(0)^{\odot2}-v(0)^{\odot2}\right|}{2}\text{sinh}\left(2\xi(t)+\log\left|\frac{u(0)+v(0)}{u(0)-v(0)}\right|\right),
\end{equation}
where $\xi:t\mapsto -\int_0^t\nabla\mathcal{L}(\theta(s))ds$ is the solution of
\begin{equation*}
    \dot{\xi}(t)+\nabla\mathcal{L}(\theta(t))=0.
\end{equation*}

\paragraph{Deriving the implicit bias} Suppose now that $\lim\limits_{t\rightarrow+\infty}\theta(t)=\theta_\infty$ such that $X\theta_\infty=y$, and that the loss function $\mathcal{L}$ can be written as $\mathcal{L}(\theta)=\ell(X\theta-y)$ for any $\theta\in\R^d$ with $\ell:\R^n\rightarrow\R$. Proving that the trajectories generated by \eqref{eq:GF_DL_1} are implicitly biased towards a potential $\mathcal{Q}$ consists in showing that:
\begin{equation}\label{eq:IB_vanilla}
    \theta_\infty=\arg\min\limits_{X\theta=y}\mathcal{Q}(\theta).
\end{equation}
In order to achieve this, \cite{woodworth2020kernel} notice that $\theta_\infty=\Psi(X^T\nu)$ (where $\Psi$ is defined in \eqref{eq:Psi2}) for some $\nu\in\R^n$. Indeed, 
\begin{equation*}
    \lim\limits_{t\rightarrow+\infty}\xi(t)=-\int_0^{+\infty}\nabla\mathcal{L}(\theta(s))ds = -X^T\underbrace{\int_0^{+\infty}\nabla\ell(X\theta(s)-y)ds}_{:=-\nu}.
\end{equation*}
This property is crucial since by writing the KKT conditions associated to the problem described in \eqref{eq:IB_vanilla}, we obtain:
\begin{equation}
    \left\{\begin{gathered}
        X\theta_\infty=y\\
        \exists\omega\in\R^n,~\nabla\mathcal{Q}(\theta_\infty)=X^T\omega
    \end{gathered}\right.
\end{equation}
The first condition is satisfied by assumption. Defining $\mathcal{Q}$ such that $\nabla \mathcal{Q}(\theta)=\Psi^{-1}(\theta)$ for any $\theta$, we get that $\nabla \mathcal{Q}(\theta_\infty)=\Psi^{-1}(\theta_\infty)=X^T\nu$, demonstrating that \eqref{eq:IB_vanilla} holds. In this case, we have
\begin{equation}\label{eq:entropy_DLN}
    \mathcal{Q}(\theta) = \frac{1}{2}\sum_{i=1}^d\left(2\theta_i\text{arcsinh}\left(\frac{2\theta_i}{\Delta_0}\right)-\sqrt{4\theta_i^2+\Delta_0^2}+\Delta_0\right)-\frac{1}{2}\left\langle\log\left|\frac{u(0)+v(0)}{u(0)-v(0)}\right|,\theta\right\rangle,
\end{equation}
where $\Delta_0 = \left|u(0)^{\odot2}-v(0)^{\odot2}\right|$.

\paragraph{Revealing a Mirror Flow} The above strategy provides insights on the implicit bias induced by reparameterizing the problem and also allows one to reveal a more general property on the trajectory of $\theta$.\\
The function $\mathcal{Q}$ defined in \eqref{eq:entropy_DLN} ensures that for any $\theta$, $\nabla \mathcal{Q}(\theta)=\Psi^{-1}(\theta)$. In particular, for any $t\geqslant0$:
\begin{equation}
    \nabla \mathcal{Q}(\theta(t))=\Psi^{-1}(\theta(t))=\xi(t),
\end{equation}
according to \eqref{eq:Psi2}. As mentionned before, $\dot\xi(t)+\nabla\mathcal{L}(\theta(t))=0$ and thus:
\begin{equation}
    \frac{d\nabla\mathcal{Q}(\theta(t))}{dt}+\nabla\mathcal{L}(\theta(t))=0.
\end{equation}
This shows that $\theta$ is the solution of a Mirror Flow dynamic with mirror map $\mathcal{Q}$.

\subsection{Detailed computations for Diagonal Linear Networks}\label{app:DLN}
Recall that we focus on the problem:
\begin{equation*}
    \min_{\theta\in\R^d}\mathcal{L}(\theta),
\end{equation*}
for some loss function $\mathcal{L}:\R^{d}\rightarrow\R$, and we consider the reparameterization $\theta=u\odot v$. By applying a Gradient Flow to the hyperparameters $u$ and $v$ we obtain:
\begin{equation}\label{eq:GF_DL_1}
    \left\{\begin{aligned}
        &\dot{u}(t)+v(t)\odot\nabla\mathcal{L}(\theta(t))=0\\
        &\dot{v}(t)+u(t)\odot\nabla\mathcal{L}(\theta(t))=0.
    \end{aligned}\right.
\end{equation}
It is then trivial to observe that $(z_+,z_-)$ defined as $(u+v,u-v)$ is a solution of:
\begin{equation}
    \left\{\begin{aligned}
        &\dot{z}_+(t)+z_+(t)\odot\nabla\mathcal{L}(\theta(t))=0\\
        &\dot{z}_-(t)-z_-(t)\odot\nabla\mathcal{L}(\theta(t))=0
    \end{aligned}\right.,
\end{equation}
which guarantees that for any $t\geqslant0$,
\begin{equation}
    \left\{\begin{aligned}
        &z_+(t)=z_+(0)e^{\xi(t)}\\\
        &z_-(t)=z_-(0)e^{-\xi(t)}
    \end{aligned}\right.,
\end{equation}
where $\xi:t\mapsto -\int_0^t\nabla\mathcal{L}(\theta(s))ds$. Note by defining $\xi$ in this way, it holds that
\begin{equation*}
    \dot\xi(t)+\nabla\mathcal{L}(\theta(t)))=0,\quad \xi(0)=0.
\end{equation*}
It is done by observing that $\theta(t)=\frac{z_+(t)^{\odot2}-z_-(t)^{\odot2}}{4}$, implying that
\begin{equation}
    \begin{aligned}
        \theta(t)&=\frac{z_+(0)^{\odot2}e^{2\xi(t)}-z_-(0)^{\odot2}e^{-2\xi(t)}}{4}\\
        &=\frac{|z_+(0)\odot z_-(0)|}{2}\text{sinh}\left(2\xi(t)+\log\left|\frac{z_+(0)}{z_-(0)}\right|\right).
    \end{aligned}\label{eq:Psi}
\end{equation}

\subsection{Study of a Deep Redundant Linear Network}\label{app:red}

In this section, we emphasize that the strategy stated in Section \ref{sec:DLN} is general enough to be applied to other similar parameterizations. We briefly analyze the implicit features of a Deep Redundant Linear Network defined through similar computations. The stated results can be seen as a simplified version of that obtained in \cite{woodworth2020kernel,moroshko2020implicit,chou2023more} in the $L$-layer case.

We consider the parameterization $\theta=u^{\odot L}$ where the number of layers $L\in\N$ is strictly greater than $2$. We also suppose that the loss function $\mathcal{L}$ can be written as $\mathcal{L}(\theta)=\ell(X\theta-y)$ for any $\theta\in\R^d$, with $\ell:\R^n\rightarrow\R$ which reaches its minimum at $0$.

By applying a Gradient Flow to $\mathcal{L}(\theta)$ according to the vector of hyperparameters $u$, it holds that
\begin{equation}
    \dot{u}(t)+Lu(t)^{\odot(L-1)}\odot\nabla\mathcal{L}(\theta(t))=0.
\end{equation}
Supposing that for any $i\in\R^d$, $u_i(t)$ is different from $0$ for any $t\geqslant 0$ (which can be done by enforcing positivity of $X$, $y$ and the initialization), we can write that
\begin{equation}
    u(t)=\left(u(0)^{\odot-(L-2)}-L(L-2)\xi(t)\right)^{\odot-\frac{1}{L-2}},
\end{equation}
where $\xi:t\mapsto-\int_0^t\nabla\mathcal{L}(\theta(s))ds$. The original variable $\theta$ then satisfies:
\begin{equation}
    \theta(t)=\left(u(0)^{\odot-(L-2)}-L(L-2)\xi(t)\right)^{\odot-\frac{L}{L-2}}=:\Psi\left(L(L-2\right)\xi(t)).
\end{equation}
The inverse of $\Psi$ is trivial to compute as for any $\theta$, $\Psi^{-1}(\theta)=u(0)^{\odot-(L-2)}-\theta^{\odot-\left(1-\frac{2}{L}\right)}$. Then, simple computations show that the entropy function $\mathcal{Q}$ defined as:
\begin{equation}
    \mathcal{Q}(\theta)=\langle u(0)^{\odot(L-2)},\theta\rangle-\frac{L}{2}\left\langle\mathbf{1},\theta^{\odot\frac{2}{L}}\right\rangle,
\end{equation}
satisfies $\nabla \mathcal{Q}(\theta)=\Psi^{-1}(\theta)$ for any $\theta$. Then, by applying the same arguments as in the previous section, we get that:
\begin{equation}
    \frac{d\nabla\mathcal{Q}(\theta(t))}{dt}+\nabla\mathcal{L}(\theta(t))=0,
\end{equation}
and, if the trajectory converges to $\theta_\infty$ a solution of the problem, i.e. $X\theta_\infty=y$, then
\begin{equation}
    \theta_\infty=\arg\min\limits_{X\theta=y}\mathcal{Q}(\theta).
\end{equation}

\subsection{Detailed proof of Theorem \ref{thm:1}}\label{app:thm1}
The proof of Theorem \ref{thm:1} follows the steps described in Section \ref{sec:thm1}.\\
$\cdot$ We start by defining $\mathcal{M}$ in the following way:
\begin{equation}\label{eq:manifold_M}
    \mathcal{M}=\left\{w\in\R^{L\times d},~\forall i\in[d],~\#\left\{j\in[L],~w_{(i-1)L+(j-1)}=0\right\}\leqslant1\right\}.
\end{equation}
We can easily show that this set is a smooth submanifold of $\R^{L\times d}$. Let us rewrite $\mathcal{M}$ in the following way:
\begin{equation*}
    \mathcal{M}=\bigcap_{i\in[d]}\left\{w\in\R^{L\times d},~\#\left\{j\in[L],~w_{(i-1)L+(j-1)}=0\right\}\leqslant1\right\}.
\end{equation*}
As a consequence, the complement set of $\mathcal{M}$ in $\R^{L\times d}$ denoted $\mathcal{M}^C$ satisfies:
\begin{equation*}
    \begin{aligned}
        \mathcal{M}^C&=\bigcup_{i\in[d]}\left\{w\in\R^{L\times d},~\exists (j,k)\in[L]^2,~j\neq k,~w_{(i-1)L+(j-1)}=w_{(i-1)L+(k-1)}=0\right\}\\
        &=\bigcup_{i\in[d]}\bigcup_{(j,k)\in[L]^2,~j\neq k}\left\{w\in\R^{L\times d},~w_{(i-1)L+(j-1)}=w_{(i-1)L+(k-1)}=0\right\}.
    \end{aligned}
\end{equation*}
We can deduce that $\mathcal{M}^C$ is a closed set and that $\mathcal{M}$ is a smooth submanifold of $\R^{L\times d}$.\\
One can notice that if $\theta$ is parameterized by $\left(u^j\right)_{j\in[L]}$ satisfying \ref{eq:assumption1}, then the corresponding $\vu(0)$ belongs to $\mathcal{M}$.\\
$\cdot$ We now prove that $G:\mathcal{M}\rightarrow\R^d$ is a commuting parameterization. Recall that it is said to be commuting on $\mathcal{M}$ if for any $(i_1,i_2)\in[d]\times[d]$ and $w\in \mathcal{M}$,
\begin{equation*}
    \nabla^2G_{i_1}(w)\nabla G_{i_2}(w)-\nabla^2G_{i_2}(w)\nabla G_{i_1}(w)=0.
\end{equation*}
We have that for any $w\in\mathcal{M}$ and $i\in[d]$,
\begin{equation}\label{eq:grad_G}
    \forall j\in [Ld],~\left(\nabla G_i(w)\right)_j=\left\{\begin{aligned}\prod_{k=(i-1)L+1,~k\neq j}^{iL}&w_k\quad\text{if}~j\in\llbracket(i-1)L+1,iL\rrbracket,\\&0\quad\text{else.}\end{aligned}\right.
\end{equation}
Thus, for any $i\in[d]$, $\left(\nabla^2G_i(w)\right)_{jk}=0$ as long as $j\notin\llbracket(i-1)L+1,iL\rrbracket$ or $k\notin\llbracket(i-1)L+1,iL\rrbracket$. As a consequence, for any $(i_1,i_2)\in[d]\times[d]$ such that $i_1\neq i_2$, $\nabla^2G_{i_1}(w)\nabla G_{i_2}(w)=0$. We can deduce that $G$ is a commuting parameterization on $\mathcal{M}$.\\
$\cdot$ It is also required that $G$ is a regular parameterization on $\mathcal{M}$. More specifically, we need to show that $J_G(w)$ is of rank $d$ for all $w \in \mathcal{M}$.\\
We can easily obtain $J_G(w)$ from \eqref{eq:grad_G} and one can observe that it has the following structure:
\begin{equation*}
    J_G(w)=\left[\begin{array}{c c c c c c c c c c} 
       K^1_1 & \cdots & K^1_L & 0 & \cdots & 0 & \cdots & 0 & \cdots & 0\\ 
       0 & \cdots & 0 & K^2_1 & \cdots & K^2_L & \cdots & 0 & \cdots & 0 \\ 
       \vdots & \ddots & \vdots & \vdots & \ddots & \vdots &  \ddots & \vdots & \ddots & \vdots \\
       0 & \cdots & 0 & 0 & \cdots & 0 & \cdots & K^d_1 & \cdots & K^d_L \\ 
    \end{array} \right],
\end{equation*}
where for any $(i,j)\in[d]\times[L]$, $$K^i_j=\prod_{k=(i-1)L+1,~k\neq j+L(i-1)}^{iL}w_k.$$
Hence, if for any $i\in[d]$ there exists $j\in[L]$ such that $K^i_j\neq0$, then $J_G(w)$ is of rank $d$ and $G$ is a regular parameterization. Following the definition of $\mathcal{M}$, this condition is satisfied for all $w\in\mathcal{M}$ which ensures that $G$ is regular on $\mathcal{M}$.\\
$\cdot$ We finally need to check that \cite[Assumption~3.5]{li2022implicit} is satisfied by $G$, i.e that the domain of the flows induced by its gradient vector fields is pairwise symmetric. We refer the curious reader to the reference for more details, but we simply notice that since $\nabla G_j(w)$ and $\nabla G_k(w)$ have non zero entries on disjoint sets of components (if $j\neq k$), this assumption is directly satisfied.

From this point, we apply \cite[Lemma 4.8 and Theorem 4.9]{li2022implicit} and conclude the first part.
{\color{black} To prove \eqref{eq:mirror_rate}, note that
\begin{align*}
\frac{d}{dt}\mathcal{L}(\theta(t))&=\langle \nabla\mathcal{L}(\theta(t)),\dot\theta(t)\rangle\\
&=-\langle \nabla^2Q(\theta(t))\dot{\theta}(t),\dot\theta(t)\rangle\\
& \leq 0,
\end{align*}
which implies that $\mathcal{L}(\theta(t))$ is non increasing.
Moreover,
\begin{align*}
\frac{d}{dt} D_Q(\theta^*,\theta) &= \frac{d}{dt}\Big[Q(\theta^*)-Q(\theta(t))-\langle \nabla Q(\theta(t)),\theta^*-\theta(t)\rangle\Big] \\
&=-\langle \nabla Q(\theta(t)),\dot{\theta}\rangle+\langle \nabla Q(\theta(t)),\dot{\theta}\rangle-\langle \frac{d}{dt}\nabla Q(\theta(t)),\theta^*-\theta(t)\rangle\\
&=\langle \nabla \mathcal{L}(\theta(t)),\theta^*-\theta(t)\rangle\\
&\leq -(\mathcal{L}(\theta(t))-\mathcal{L}^*).
\end{align*}
Therefore, $(\mathcal{L}(\theta(t))-\mathcal{L}^*)$ is decreasing and integrable, and
\[
D_Q(\theta^*,\theta(t)) +\int_0^t (\mathcal{L}(\theta(s))-\mathcal{L}^*)ds \leq D_Q(\theta^*,\theta(0)).
\]
The statement then follows. }
\end{document}